\documentclass[sigconf,letterpaper]{acmart}
\pdfpagewidth=\paperwidth
\pdfpageheight=\paperheight
\usepackage[utf8]{inputenc}
\usepackage{natbib}
\usepackage{subfigure}

\graphicspath{ {images/} }
\begin{document}

%\title{Evolving Playable Hearthstone Decks With MAP-Elites}
%\title{Curating Hearthstone Deck Spaces\\Using MAP-Elites With Sliding Boundaries}
\title{Mapping Hearthstone Deck Spaces\\through MAP-Elites with Sliding Boundaries}

%\author{Matt, Scott, Lisa, Fernando, Julian, Amy}

\author{Matthew C. Fontaine}
\affiliation{%
  \institution{Independent Researcher}
}
\email{tehqin@gmail.com}

\author{Scott Lee}
\affiliation{%
  \institution{Independent Researcher}
}
\email{randomperson2727@gmail.com}

\author{L. B. Soros}
\affiliation{%
  \institution{Champlain College}
  \city{Burlington} 
  \state{VT} 
}
\email{lisa.soros@gmail.com}

\author{Fernando De Mesentier Silva}
\affiliation{%
  \institution{Independent Researcher}
}
\email{fms2005@gmail.com}

\author{Julian Togelius}
\affiliation{%
  \institution{Tandon School of Engineering\\New York University}
  \city{New York City} 
  \state{NY} 
}
\email{julian@togelius.com}

\author{Amy K. Hoover}
\affiliation{%
  \institution{Ying Wu College of Computing\\New Jersey Institute of Technology}
  \city{Newark} 
  \state{NJ} 
}
\email{ahoover@njit.edu}

\renewcommand{\shortauthors}{M. Fontaine et al.}

\date{}

% TODO
% Not sure whether to introduce the abbreviation for map-elites here. Currently it is used but not ever made clear what it stands for

\begin{abstract}
Quality diversity (QD) algorithms such as MAP-Elites have emerged as a powerful alternative to traditional single-objective optimization methods. They were initially applied  to evolutionary robotics problems such as locomotion and maze navigation, but have yet to see widespread application. We argue that these algorithms are perfectly suited to the rich domain of video games, which contains many relevant problems with a multitude of successful strategies and often also multiple dimensions along which solutions can vary.

This paper introduces a novel modification of the MAP-Elites algorithm called \emph{MAP-Elites with Sliding Boundaries} (MESB) and applies it to the design and rebalancing of Hearthstone, a popular collectible card game chosen for its number of multidimensional behavior features relevant to particular styles of play. To avoid overpopulating cells with conflated behaviors, MESB slides the boundaries of cells based on the distribution of evolved individuals. Experiments in this paper demonstrate the performance of MESB in Hearthstone. Results suggest MESB finds diverse ways of playing the game well along the selected behavioral dimensions. Further analysis of the evolved strategies reveals common patterns that recur across behavioral dimensions and explores how MESB can help rebalance the game. %Thus, the paper introduces both a modification of the MAP-Elites algorithm and a new use case of QD algorithms for understanding and improving game design and balancing.
%the boundaries of behavior features to adapt to variations in the distribution of interesting solutions in the map of elites generated by the algorithm.

\end{abstract}

%\ccsdesc[500]{Computer systems organization~Embedded systems}


\maketitle

% TODO in intro (maybe also abstract)
% This is a well written paper though a bit confused -- it is not clear what the exact goal of this work is, or what the ultimate contributions are, and no research questions as proposed as such.

% TODO also intro
% After explaining how good current novelty search variants are, why then introduce a new algorithm? What is lacking in existing approaches that this new approach adds?

\section{Introduction}

% What big problem does the field currently face, and why is it important/interesting/challenging?
Quality diversity (QD) algorithms have recently emerged as a powerful alternative to traditional single-objective optimization methods \citep{pugh:frontiers16}. Because of their ability to discover multiple and diverse optima in a search space, they are well-suited for domains with many types of viable solutions. In comparison to single-objective optimization methods, QD algorithms may better approximate the variety of strategies that humans develop to navigate complex real-world environments.

Despite their potential impact, QD algorithms are predominately explored in only a fraction of the possible domains that may benefit from them, including problems in traditional evolutionary robotics such as locomotion \citep{cully:nature15} and maze navigation \citep{pugh:frontiers16}. However, games offer a potentially fruitful avenue for QD research because of the multitude of possible strategies that can result in a success or win condition, as well as the multiple dimensions along which game content can and sometimes should vary~\cite{khalifa2018talakat}. Exploring domains beyond evolutionary robotics is critical for understanding both the strengths and limitations of this new class of algorithms.

% TODO paragraph below
% "collectible card game" What does this mean exactly?

Hearthstone~\cite{hearthstone:web181}
%\footnote{\textcopyright2014, Blizzard Entertainment, Inc.}
is a popular collectible card game that presents a variety of AI-based challenges, including developing strategies for gameplaying and deckbuilding. This paper adapts the canonical MAP-Elites (short for \emph{Multi-dimensional Archive of Phenotypic Elites}) QD algorithm to generate decks in Hearthstone, where the main challenge is not to find a gameplaying or problem-solving strategy (or set of strategies), but instead to evolve a deck, which can be seen as a toolbox around which a human player can construct winning strategies. In this way, Hearthstone presents a fundamentally new \emph{kind} of domain for QD algorithms, which has previously been applied to search for controllers or strategies directly. Additionally, applying a QD algorithm to games like Hearthstone presents a novel challenge because viable decks must be able to adapt to an opponent actively and antagonistically changing the environment (herein interpreted as game state). 

% What is the main contribution of this paper?
The principal contributions of this paper are a new modification of the MAP-Elites algorithm (MESB), a new application of this modification of MAP-Elites to generating decks in Hearthstone, and several results concerning the availability of good decks in the basic and classic set of cards in Hearthstone. Called MAP-Elites with Sliding Boundaries (MESB), this new modification of MAP-Elites introduces sliding boundaries, which allows for better handling of unequal distribution of promising solutions. While there are a few earlier examples of QD \citep{khalifa2018talakat} and constrained search \citep{gravina:cig16} algorithms for content generation in games, here we not only demonstrate the viability of MAP-Elites (and QD algorithms in general) for generative deck design, but we also show how analyzing sets of diverse evolved decks can provide insights into the dynamics of the game and concrete suggestions for how to rebalance a real and widely-played collectible card game.

The following section provides a brief overview of both the computational design and analysis of collectible card games and quality diversity algorithms. Our approach, including the novel MAP-Elites modification MESB and the Hearthstone simulator is then presented. Next comes a series of experiments where we generate sets of decks under various conditions, and analyze these sets by mining frequent patterns to investigate dominant cards that recur in evolved decks that recur despite the diversity maintenance mechanism inherent in MAP-Elites. The results of a rebalancing experiment wherein the insights gathered from the analysis of evolved decks are used to change the dynamics of the game. Finally, a discussion of the results and our future work is described.

\section{Background}

This section first describes deckbuilding in Hearthstone, then reviews automated approaches to deckbuilding and playtesting before describing quality diversity and the MAP-Elites algorithm.

\subsection{Hearthstone}

Published by Blizzard, Hearthstone is an adversarial, online collectible card game, where two players take turns placing cards on the digital board shared between them. When a Hearthstone game begins, each player has 30 health points and it ends when one player's health is reduced to zero. Players are represented by one of nine possible hero classes, each with a unique hero ability that can be played at most once per turn and a subset of cards only playable by the class. While there are over 1900 possible cards to collect, at the start of each turn players draw one card to add to their hand from a subset of 30 that they have pre-selected for their decks. Deckbuilding and deciding when and how to play cards in the deck are two distinct strategy challenges. This paper explores strategies for building decks, leaving the gameplay strategies constant for each player.

%Playing Hearthstone competitively requires two distinct strategy components, both 1) real-time playing of these cards, and 2) selecting which cards will be in the deck. While selecting which cards to play and win is an interesting challenge, inherent to these approaches is building a 
%These cards are the main mechanism for real-time gameplay as they dictate when and how players can attack each other. Playing Hearthstone competitively requires two distinct strategy components: 1) composing a deck (i.e. choosing exactly 30 of over 1400 possible cards accessible to the player in a given game), and 2) playing it strategically.

While Blizzard has so far released thirteen sets of cards, when initially introduced in 2014 there were two: basic and classic, which together contain 171 playable cards for each class. Cards can be added to a deck at most twice, except for special \emph{legendary} cards that can be included only once. Because there are approximately $1.42 \times {10}^{35}$ decks that can be composed for a given class with these two sets of cards, finding quality decks within this space is a significant challenge. While future experiments will explore deckbuilding with cards from different sets, experiments in this paper focus on the initial two.

%Decks each containing 30 cards are constructed using cards playable by the chosen hero. All cards can be included in a deck at most twice, except for special \emph{legendary} cards that can be included only once. For a given class, approximately $1.42 \times {10}^{35}$ decks exist using cards from the basic and classic sets.

\begin{figure}
\centering
{
    \includegraphics[scale=0.4]{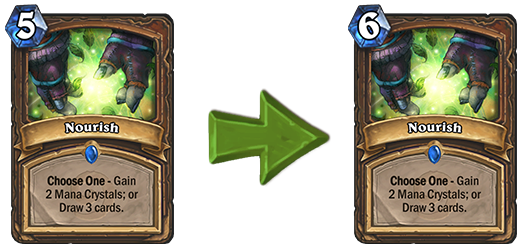}
}
\caption{\emph{Card Nerf.} Because this card was too powerful, \emph{Nourish} was nerfed to balance the game in December 2018 from costing five mana to its current cost of six. 
\label{fig:nerfex}}
\end{figure}

Because of the complexity of deckspace, it is difficult to balance. However, balance is necessary to ensure players can win with a variety of decks, heroes, and gameplay strategies. Blizzard regularly edits the properties of popular cards to increase (i.e. \emph{buff}) or decrease (\emph{nerf}) their power and popularity in the Hearthstone community. Figure~\ref{fig:nerfex}) shows an example nerf to the card \emph{Nourish}, which was nerfed by increasing the mana cost. It is difficult to determine \emph{a priori} which nerfs and buffs will balance the game, so Blizzard regularly relies on player data to make these determinations.

Because of the number of possible deckbuilding and gameplay strategies, it is important that computational tools for balancing the game are capable of simulating a diversity of strategies that reflect the variety of real-world playstyles. For example, one popular gameplay strategy is an aggressive \emph{aggro} approach where players focus on dealing damage to the opponent as quickly as possible. However, a \emph{control} strategy instead focuses on maintaining board control and dealing damage to the opponent only once control is established. There are other popular strategies like \emph{ramp}, \emph{midrange}, \emph{one turn kill}, \emph{combo}, and \emph{fatigue}, but experiments in this paper consider only the first two as they are the most straightforward to implement and therefore more likely to simulate human play. Regardless, to play well it is important that a deckbuilding strategy complements the gameplay strategy. 

\subsection{SabberStone Simulator}

% Remember to describe how our AI works either here or in the approach section

SabberStone\footnote{\url{https://github.com/HearthSim/SabberStone}} is a Hearthstone simulator that implements the rules of the game and acts directly on the card definitions provided by Blizzard. The simulation includes an AI player that implements a turn local strategy. Action sequences are randomly generated and then evaluated with a basic game tree search algorithm where game states are hashed such that each game state is evaluated at most once. When an action sequence reaches the end of a turn, the game state is evaluated by a heuristic for playing the game well. Different heuristics can mimic specific play styles, like the aggro and control heuristics previously described. 

\subsection{Quality Diversity and MAP-Elites} 
Quality diversity (QD) algorithms, sometimes referred to as \emph{illumination algorithms}, are inspired by the ability of evolution in nature to discover many niches and many different viable strategies for survival and reproduction \cite{pugh:frontiers16}. This paradigm differs from traditional evolutionary computation, which optimizes towards a single objective, but also from multiobjective evolutionary optimization, where the trade-offs between multiple objectives are explored. 

Novelty search \cite{lehman:alife08}, which replaces the standard fitness gradient with a reward for finding individuals that are simply different from anything found previously, is an important predecessor of modern QD algorithms. This strategy is beneficial because finding high-performing regions of so-called \emph{deceptive} search spaces may require traversing intermediate ``stepping stones'' found in low-fitness regions. Since the advent of novelty search, the field of evolutionary computation has seen a focused interest in algorithms exploring beyond the fitness-only approach, for instance by combining novelty with multi-objective search \cite{mouret:evorob11}. Note that typical multi-objective optimization algorithms do not necessarily have this stepping stone collection property. At the time of this writing, the most popular QD algorithms are MAP-Elites \citep{cully:nature15} and novelty search with local competition (often abbreviated as NSLC)\cite{lehman:gecco11}. These algorithms differ from vanilla novelty search because they each incorporate a performance-based \emph{quality} measure combined with an incentive for novelty or diversity. MAP-Elites in particular has been popular recently due to its relative simplicity and strong performance on evolutionary robotics domains such as many-legged locomotion. Inspired by the idea of collecting a behavioral repertoire \citep{cully:gecco13, cully:ec16}, MAP-Elites imposes a discretized grid over a continuous behavior space and then collects the highest-performing individuals within each grid cell. In this way, the algorithm maintains a diverse set of phenotypes from which to generate new populations. The algorithm's ability to thereby find \emph{many} solutions to a given problem makes it particularly applicable to Hearthstone.
%for which there exist practically inexhaustible viable strategies and deck compositions.

%\cite{vassiliades:ec18}

\subsection{Automated Deckbuilding and Playtesting} % 

As far as we know, no previous work has applied QD algorithms to deckbuilding and playtesting. This section therefore reviews non-QD approaches (including some non-evolutionary methods) to these problems.

%not hearthstone

Evolutionary methods can help explore the design spaces of games by finding hidden aspects of these spaces, potentially revealing novel insights. For example, \citet{togelius2008experiment,browne2010evolutionary} evolve new games predicted to be interesting for human players. Yavalath is an example of such a game that was commercially published and currently has a ranking of 7.2 on the popular website for ranking board games called BoardGameGeek \footnote{https://boardgamegeek.com/boardgame/33767/yavalath}
Other approaches explore the design and strategy spaces of \emph{particular} games like creating variants of Flappy Bird by tuning game parameters \cite{isaksen2015discovering}. \citet{de2016generatingblackjack} search the space of gameplay strategies for Blackjack and Poker~\cite{de2018generatingprefloppoker, de2018generatingfullpoker}. While the discovered heuristics were simple and intended to instruct novice players, they were often comparable or more successful to those describing more complex strategies. 

%~\citep{silva2018exploring, de2017ai,horn2016opening}

%Game Balancing
Games can be fundamentally rebalanced with even small changes to the rules \citep{isaksen2016characterising}, mechanics representation \citep{de2018drawing} or game content \citep{kowalski2018strategic}. However, \citet{hom2007automatic} explore automated approaches to balancing through changing game rules until agents are able to play against each other with relatively equal winrates and few draws. Through his Machinations framework, \citet{dormans2011simulating} helps designers make small rule changes early on to ensure balance throughout the design process. Alternatively, \citet{jaffe2012evaluating} present a framework to evaluate balance through comparing standard gameplaying agents to those with restricted freedom. Results indicate that such a process can help balance an educational card game. Like the previous approach, \citet{preuss2018integrated} balance games through integrating human and automated testing. While these types of changes may be necessary for game balance, they often have a large impact on the gameplay.

%We apply a multi-objective evolutionary algorithm to obtain decks that optimise objectives, e.g. win rate and average number of tricks, developed to express the fairness and the excitement of a game of top trumps. The results are compared with decks from published top trumps decks using simulation-based objectives. The possibility to generate decks better or at least as good as decks from published top trumps decks in terms of these objectives is demonstrated. Our results indicate that automatic balancing with the presented approach is feasible even for more complex games such as real-time strategy games.

 There are several approaches to evolutionary deckbuilding that this paper builds upon. \citet{volz2016demonstrating} explore evolutionary deckbuilding for the game Top Trumps that optimized different objectives with a goal of expressing fairness in the game. However, in Top Trumps all of the cards in a given pack are distributed to all of the players, whereas Hearthstone players build their own decks individually and with intention. \citet{mahlmann2012evolving} also searches for balanced card sets in Dominion through automated agents with different skill levels and evolutionary parameters. While in Hearthstone players must build their decks before playing, decks in Dominion are built through play. While AI-based approaches to playtesting have proven effective, the Hearthstone domain poses some unique challenges.  .   %\citet{krucher2015algorithmically} attempted to use artificial intelligence to modify a set of cards to create a balanced pool. 

% hearthstone

Hearthstone is a particularly challenging game because of the amount of information hidden from the player, stochasticity, and high branching factor. As a result, there are many approaches to creating AI agents to play Hearthstone~\cite{swiechowski2018improving, da2018hearthbot, zhang2017improving, santos2017monte, stiegler2017symbolic, grad2017helping, dockhorn2018predicting}.  Furthermore, there have been significant advancements in win predictions based on game state evaluation~\cite{janusz2017helping, janusz2018toward, jakubik2017evaluation}. \citet{bursztein2016legend} took a unique approach by building a predictor to determine the next card that the opponent is likely to play. Modeling gameplay strategies in Hearthstone is an open problem.

Hearthstone deckbuilding has been explored with techniques other than evolution. Chen et al. presented a deck recommendation system that makes suggestions to improve the performance on the current match-up~\cite{chen2018q}. Stiegler et al. introduced a utility metric to classify cards in relation to a deck being built~\cite{stiegler2016hearthstone}. The method adds the cards with highest utility to the deck and then proceeds to recalculate the utility of the remaining card pool. All methods output one deck as a result.  Zook et al. previously evaluated how design choices impact gameplay using a simplified version of Hearthstone for case studies~\cite{zook2015monte, zook2018learning}. Jin proposed a method for measuring card balance and consequently deck strength~\cite{jin2018proposed}, while Janusz et al. investigated card similarity based on their text embedding~\cite{janusz2018investigating}. 

Previous work on evolutionary Hearthstone deckbuilding in particular employed non-QD evolutionary algorithms to generate better starter decks by evaluating decks against a single AI opponent~\cite{bhatt2018exploring}. Garc\'ia-S\'anchez et al. for Hearthstone~\cite{garcia2016evolutionary, garcia2018automated} similarly evaluated evolved decks against a suite of opponents.
Similar approaches proved successful for the deckbuilding game Magic: The Gathering~\cite{bjorke2017deckbuilding}.

\section{Approach}

% Remind us, what is the goal of what we're doing?

%This section details the specific implementation of MAP-Elites applied to the problem of Hearthstone deck generation in this paper.

This section discusses the parameters of MESB for deckbuilding in this paper and what distinguishes the modification from the traditional MAP-Elites algorithm.

% "menagerie of adversaries" -> Fernando's note: Intransitivity concept

% Remember to describe how our AI works either here or in the Sabberstone section

\subsection{Mutation and Fitness}
Mutation is performed by replacing $k$ cards randomly from a pool of basic and classic cards that result in valid decks. The value $k$ varies geometrically where the probability of exchanging $k$ cards is given by ${Pr(X=k) = \frac{1}{2}Pr(X=k-1)}$ and ${Pr(X=1) = \frac{1}{2}}$. Geometric mutation was chosen to satisfy the maximum entropy principle. All decks can be reached using $k=1$ mutations within $30$ mutations (i.e. by swapping each card in the source deck with a differing card in the destination deck). Fitness is the sum of differences in hero health over 200 games, where a positive health difference results from victory and a negative health difference results from defeat. The health difference is used as fitness rather than the win rate so that the magnitude of victory or defeat is preserved. 

% Some notes from the background section:

%Our approach differs as it does not affect the game mechanically or its tuning, rather it explores the domain contained under the game's content, but instead for searching for strategies, we seek to identify custom settings that provide different game play experiences for human players.

%We do not alter the game mechanically, but rather try to evaluate the balance by exploring its existing content.

%We do use an AI to play the game (which we describe on section 4), it is not a high performing agent, but it does succeed in approximating two common human playing strategies.

%With the use of MAP-Elites we are able to better explore the space, and find multiple high performing decks. We then analyze the results found to evaluate diversity between the decks we encountered to discuss the options created from the cards design.

%After analyzing the most common card sets present on the decks of our MAP-Elites, we propose and evaluate the impact that modifying a small group of cards has on decks generated by our method.

\subsection{MAP-Elites with Sliding Boundaries} 

% Describe choice of behavior characterization, also justify choice
% Note that some implementations of MAP-Elites include scaling, while other do not. Make clear that this implementation does include scaling. 
% Make sure to talk about sliding boundaries and explain the implementation clearly

\begin{figure}
\centering
\includegraphics[scale=.3]{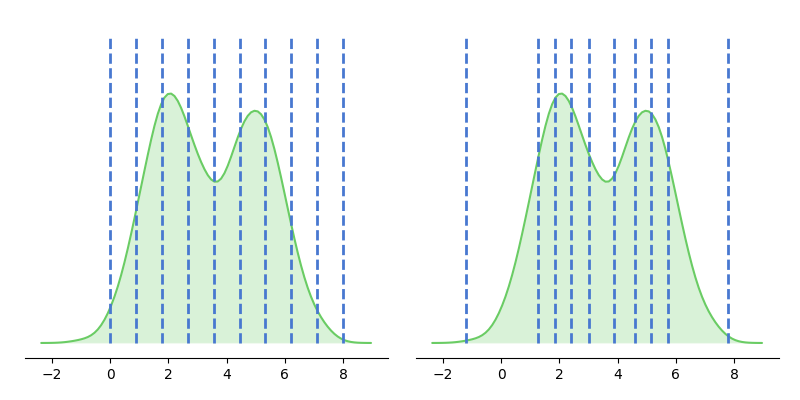}
\caption{\emph{Drawing Boundaries in the Behavior Space.} Unlike original versions of Map-Elites that draw uniform boundary lines regardless of the population density (left), MESB draws boundaries based on the number of individuals currently occupying regions of the behavior space (right).}
%The left feature distribution shows the uniform boundary lines from MAP-Elites, while the right distribution shows the percentage boundary lines utilized by MAP-Elites with sliding boundaries.}
\label{fig:sliding-ex}
\end{figure}

When considering behavior vectors, typical implementations of MAP-Elites uniformly set the boundary lines between cells, implying a bounded behavior space. Additionally, if the distribution of the behavior space is not uniform, the space illuminated by MAP-Elites will not accurately match the true distribution of the feature space. Lastly, it can be difficult to know the distribution of individuals along a feature vector \emph{a priori}. To solve these issues, the novel sliding boundaries modification is introduced. Instead of placing boundaries uniformly by feature value, the boundaries are placed at uniformly at percentage marks of the distribution (see figure~\ref{fig:sliding-ex}). To set these boundaries, a population of the last $\xi$ individuals is maintained in a queue data structure. A remap frequency $\delta$ is also specified. Every $\delta$ individuals, the boundary lines for the map are recalculated. They are recalculated by sorting the individuals along each feature and finding each individual at the corresponding percentage mark. Maintaining a sampling of the behavior space enables the estimation of the true distribution of the search space. Using binary search, queries for the proper cell can be executed in $O(d \log b)$, where $b$ is the number of boundaries and $d$ is the number of dimensions in the map. As remapping only happens periodically, the algorithm maintains good empirical performance. For all experiments $\xi=\infty$, meaning all discovered individuals are used to draw boundary lines, and $\delta = 100$, meaning the map boundaries are recalculated every $100$ individuals. 

% TODO figure below
% Figure 2 could be larger, although it's not too bad.

\subsection{MAP-Elites with Resolution Expansion}
MAP-Elites implementations can vary in certain parameters. Some implementations use a fixed number of boundaries along a given feature. Other implementations scale the number of boundaries over time. However, MESB scales linearly where the initial size of the map is $2 \times 2$ and incrementally scales the map to $20 \times 20$ at uniform time intervals (i.e. $2\times 2, 3\times 3, ..., 20\times 20$). Preliminary results showed a better performance for this scaling method than a fixed resolution of the archive.

\subsection{Behavior Vectors} 

% TODO revisit first paragraph

Knowing how the sampled distribution found by MAP-Elites varies from the complete distribution of all possible decks in the deckspace can help evaluate the sliding boundaries modification in MESB. Because it can be difficult to enumerate the distribution of all individuals in the deckspace for a set of behaviors, MESB distributes elites based on their fitness and genetic properties of the deck as behavior vectors. Such properties are statistical information that can be calculated \emph{before} evaluation, making it possible to measure complete sets of possible distributions.

%%%%%

%Many basic gameplay strategies in Hearthstone differ significantly in terms of the \emph{mana curves}, or distributions of mana costs of cards, in their corresponding decks. For instance, aggro strategies that play many cards early in the game often have few if any cards that cost over five mana (figure \ref{fig:examplemanacurve}). 
%However, both high- and-low performing decks may share the same mana curve, so actual (simulated) gameplay is still necessary in addition to mana curve analysis to determine the quality of a deck. 

%%%%%%%

In MAP-Elites (and other QD algorithms), evolved individuals are usually characterized by behaviors observed during evaluation to maintain diversity. However, meaningful diversity can be characterized by information known \emph{a priori} because gameplaying strategies in Hearthstone differ significantly based on the mana cost of cards in a deck. Such distributions of card costs are often called \emph{mana curves}, with a cost between zero and ten on the x-axis and frequency between zero and thirty on the y-axis. They are important because they often characterize the type of gameplay strategies that can be successful. For instance, aggro strategies that play many cards early in the game often have few if any cards that cost over five mana. As an approximation of this mana curve, the behavior vector for each deck contains 1) the average mana cost of all 30 cards in the deck and 2) the variance of mana costs.

% TODO add some of this info to improve figure 3 caption

In addition to being able to calculate the true distribution of average mana and mana variance, these behaviors vectors are associated with different style of play. For instance, cards with a low mana cost are typically played early in the game, while cards with a high mana cost are played later. The average mana of the deck is a rough measure on which stage of the game the deck specializes in. When deckbuilding, players must balance the mana cost of their deck focusing on more than just the early or last stages of the game. Including variance of the mana distribution measures helps measure how much the deck focused on one area of the game versus spreading out to try other mana distributions.

% \cite{decoster}
% Any differences / simplifications from actual Hearthstone?

%\subsection{Parallelism}

%Experiments were run on a high performance cluster with 500 CPU nodes running in parallel. A parallel version of MAP-Elites was implemented to take advantage of the parallel nodes. 499 workers nodes where created that evaluate a single deck for 200 games and a single coordinator node was used to run the parallel MAP-Elites. 

\section{Experiments}

% TODO figure below
% Figure 3 could be larger, although it's not too bad.

%\begin{figure}
%    \centering
%    \includegraphics[scale=.3]{evoLockMC.png}
%    \caption{\emph{Example mana curve (aggro strategy)}. Mana curves show the number of cards in a deck with specific mana costs. Successful aggro decks often have few cards that cost over five mana.
%    \label{fig:examplemanacurve}}
%\end{figure}

% TODO section below
% The authors give very little detail about the experimental setup, but then go into too much discussion in the Results without actually showing the result.

Experiments were run on a high performance cluster with 500 CPU nodes running in parallel. (A parallel version of MAP-Elites was implemented to take advantage of the parallel nodes.) The code used to run these experiments is available on GitHub as a platform called EvoSabber\footnote{https://github.com/tehqin/EvoSabber}.
Three sets of experiments were performed:

\begin{enumerate}
\item In the \textbf{MESB validation experiment} opponents play decks called starter decks, which are constructed with basic cards and available to any player. The goal is to explore whether MESB can generate a high variety of decks across different mana distributions, and whether these decks reflect mana curves appropriate for their archetype.

\item The \textbf{elite adversaries experiment} then evolves new decks against the best decks found in the map of elites from the first experiment (in other words, with the decks evolved in the MESB validation experiment as adversaries). This experiment explores the ability to evolve effective counters to known strong decks.

\item For the \textbf{deck balancing experiment}, we first perform an Apriori analysis \citep{agrawal:vldb94} on all elites in the decks of the different archetypes to identify the most commonly occurring combinations of cards, and we then explore the space of decks after altering these cards to intentionally affect balancing. In part this experiment is designed to explore whether MESB is a suitable potential tool for game designers.
\end{enumerate}

% LISA EDITING HERE

In each of these three experiments, three different deck configurations (hunter, paladin, and warlock) were paired both with aggro and control strategies. The hunter, paladin, and warlock configurations were selected in part for their reputations of supporting both aggro and control play styles well. Each deck configuration $\times$ strategy combination was evolved by MESB for 10,000 evaluations (with 200 simulated played games per evaluation) per experiment. 

The goal in each experiment is to evolve a set of high-performing set of decks that vary in terms of mana curve (which serves as a proxy for strategy type). As a reminder, fitness for the purpose of MESB elite selection in these experiments is quantified as the sum of differences in hero health over 200 games. However, winrate (the percentage of games won, regardless of health difference at end of game) is chosen to illustrate performance in these experiments because winrate determines player rankings in real-world Hearthstone games. The two-dimensional behavior space for MAP-Elites in these experiments has average mana cost of all cards in a deck on one axis and variance of these values as the other axis. Such a behavior space not only approximates a diverse strategy space but also affords easy performance visualization. Understanding how well MESB covers the space of mana curves is a good indicator of the algorithm's potential for generating nontrivially distinct decks.

\section{Results}

\subsection{MESB Validation}
%By looking at the performance across a variety of behavior constraints, MAP-Elites potentially explores a large game design space of Hearthstone. Whereas MAP-Elites necessarily parameterizes the upper and lower boundaries of the behavior vectors, MESB encourages the open exploration of all potential behavioral features. By mapping fitness across the dimensions, trends among classes and strategies emerge.

%MAP Elites by nature provides us a lower bound of the upper bound of decks, and we can use this to inform us about how these different behavior vectors correlate with evaluation as well as other vectors.
%One of the things that makes MAP-Elites especially useful for design space exploration is its ability to generate decks that span a wide variety of constraints. 

% TODO figure 6
% Fig 6 on the other hand I think should have the same axes of "average mana" and "mana variance" but are just labelled with "average" and "variance"

%\begin{figure*}
%\centering
%\includegraphics[width=0.9\textwidth]{elite_map2.png}
%\caption{\emph{Distributions of Deck Performances Evolved Against Starter Decks in the MESB Validation Experiment} 
%\label{fig:expset1-scatterplot}. Darker hues indicate positive performance (quantified by games won). Scatterplots for different deck configurations and strategies exhibit different shapes.}
%\end{figure*}

\begin{figure*}
    \centering
     \subfigure[Control Hunter]
    {
        \includegraphics[width=.32\textwidth]{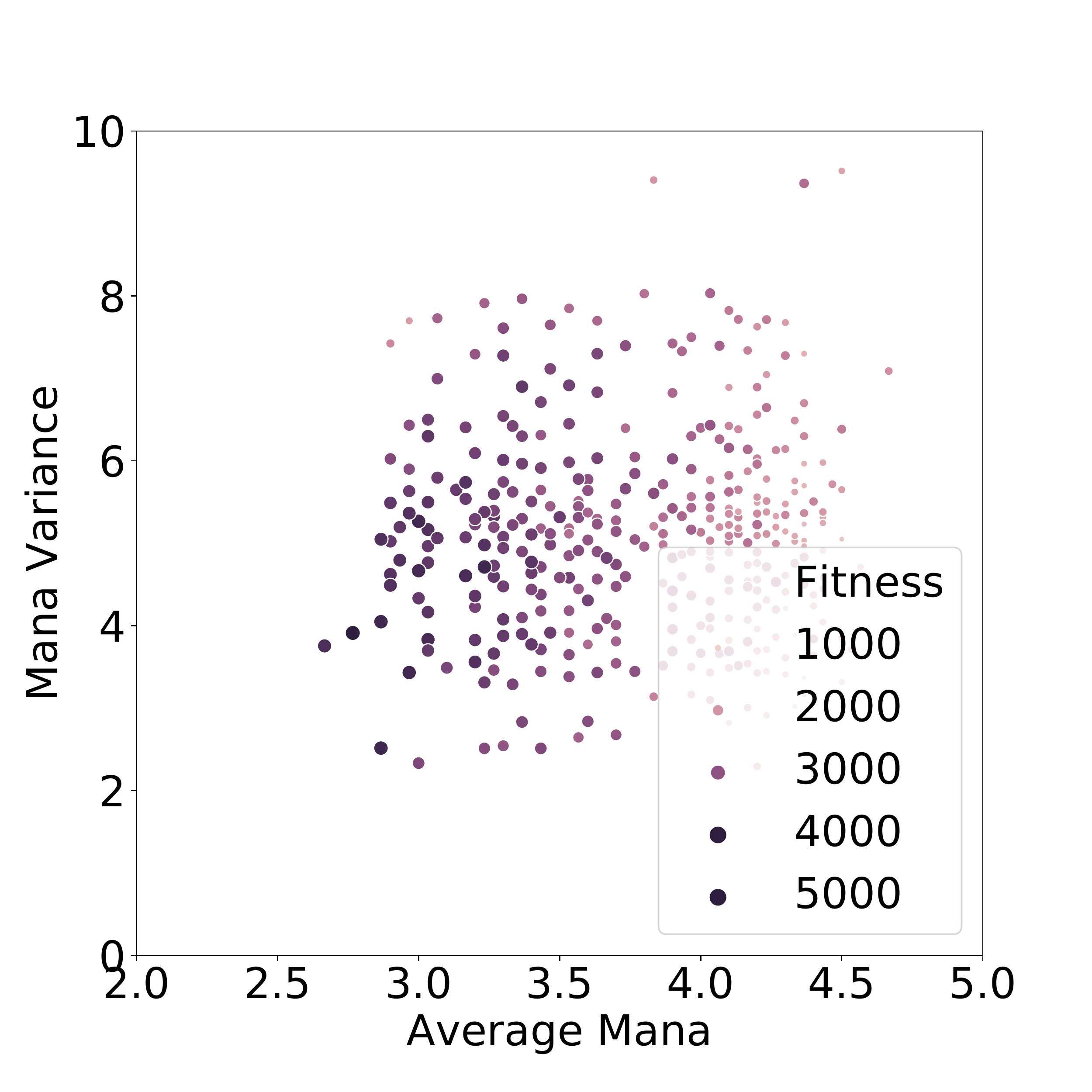}
    }
    \subfigure[Control Paladin]
    {
        \includegraphics[width=.32\textwidth]{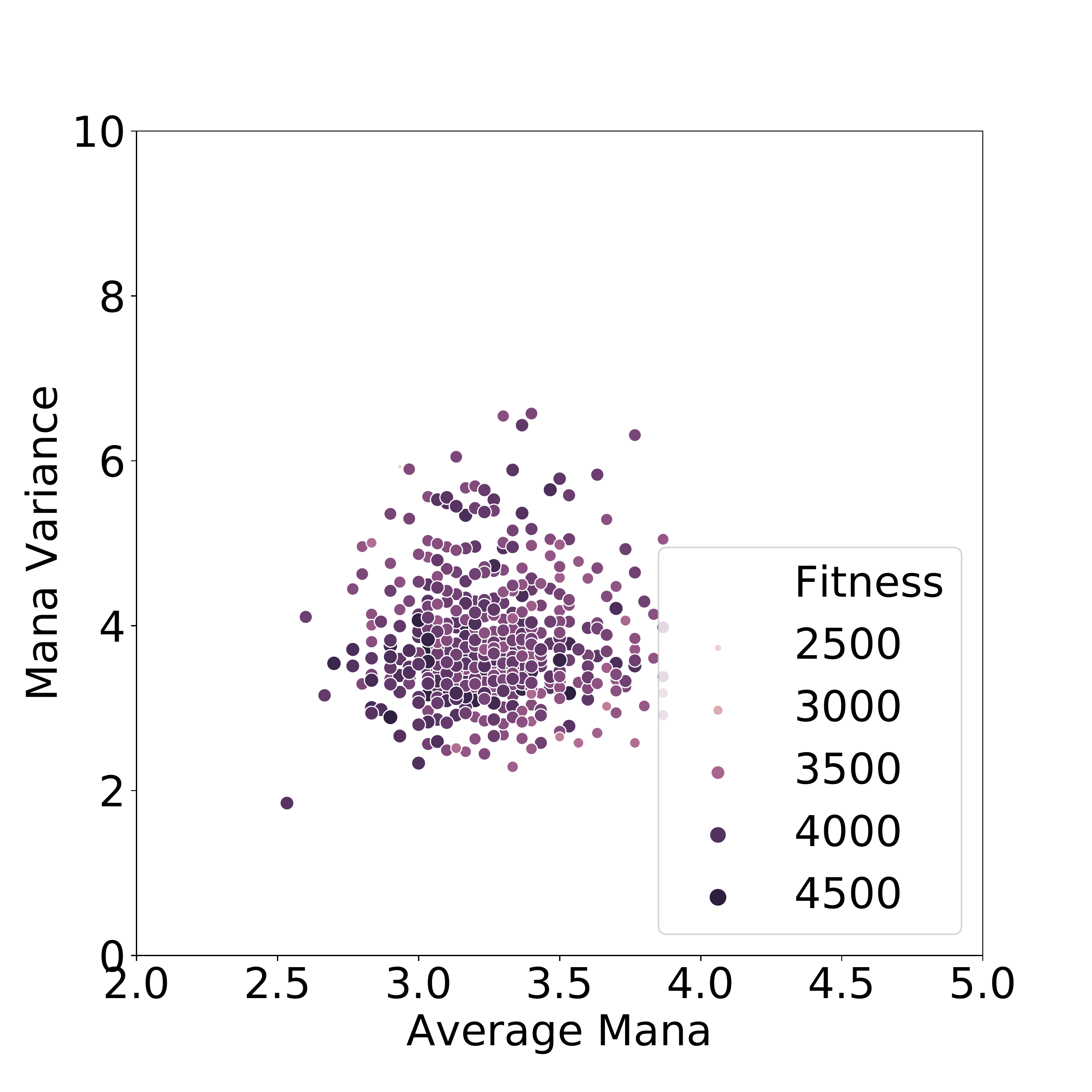}
    }
    \subfigure[Control Warlock]
    {
        \includegraphics[width=.32\textwidth]{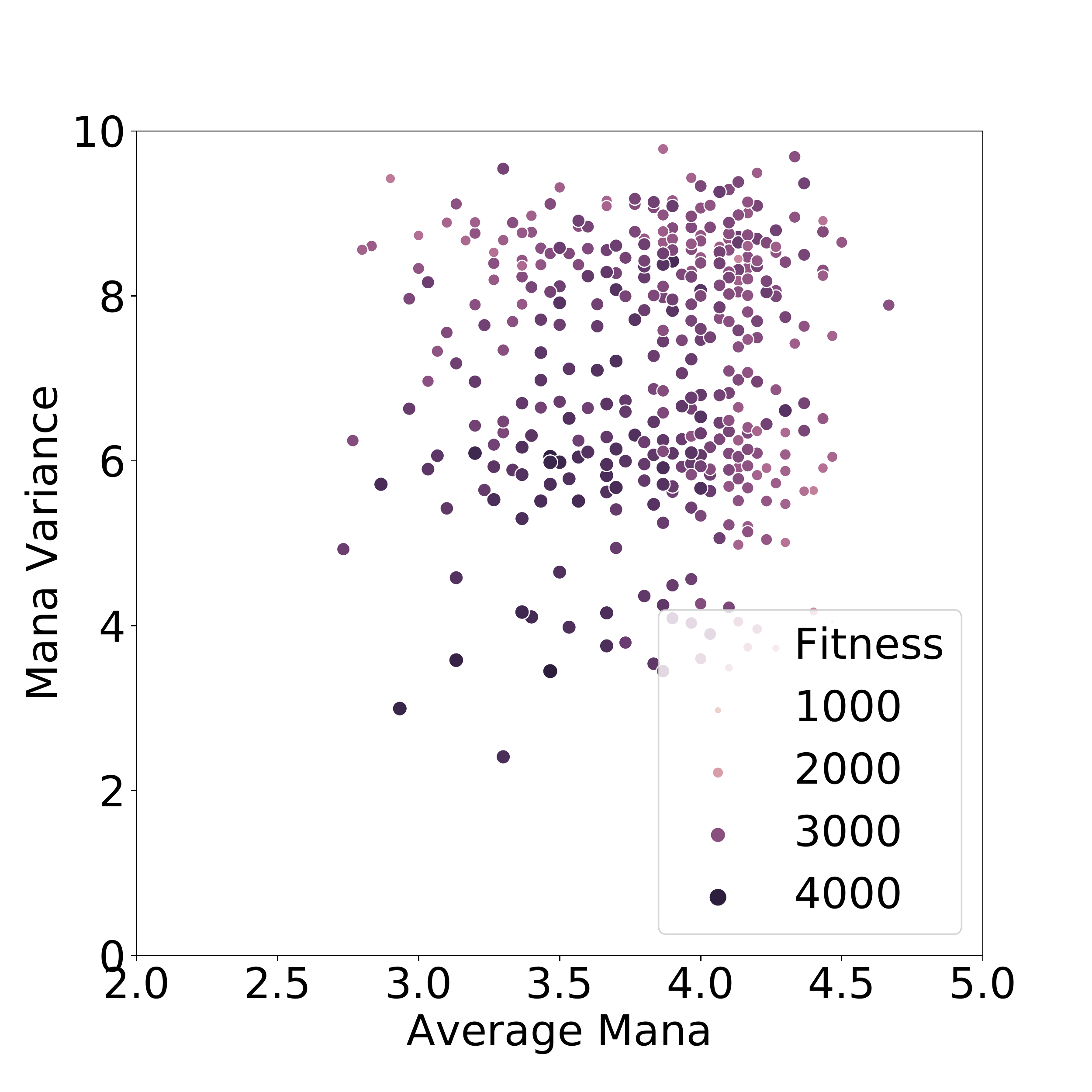}
    }
    \subfigure[Aggro Hunter]
    {
        \includegraphics[width=.32\textwidth]{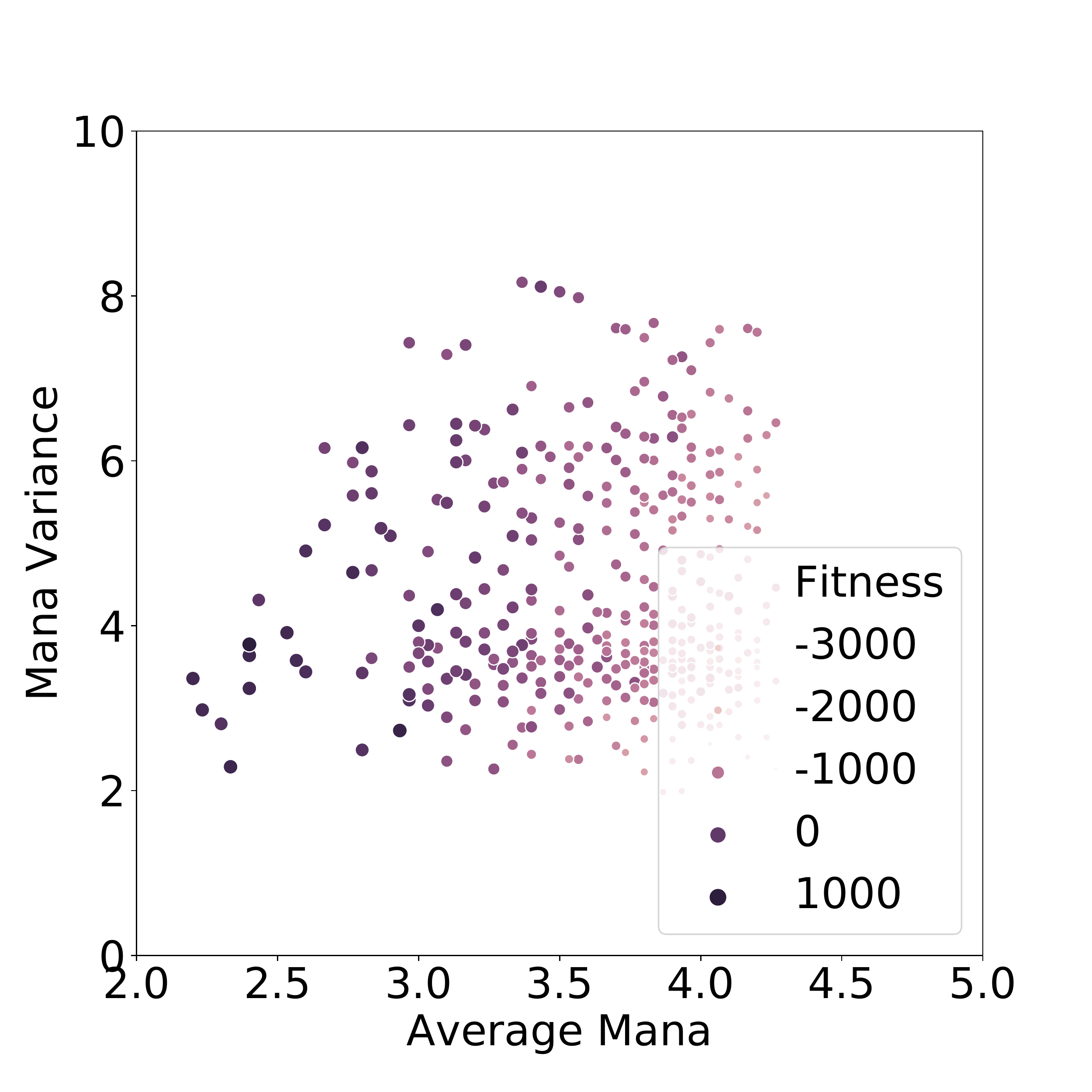}
    }
    \subfigure[Aggro Paladin]
    {
        \includegraphics[width=.32\textwidth]{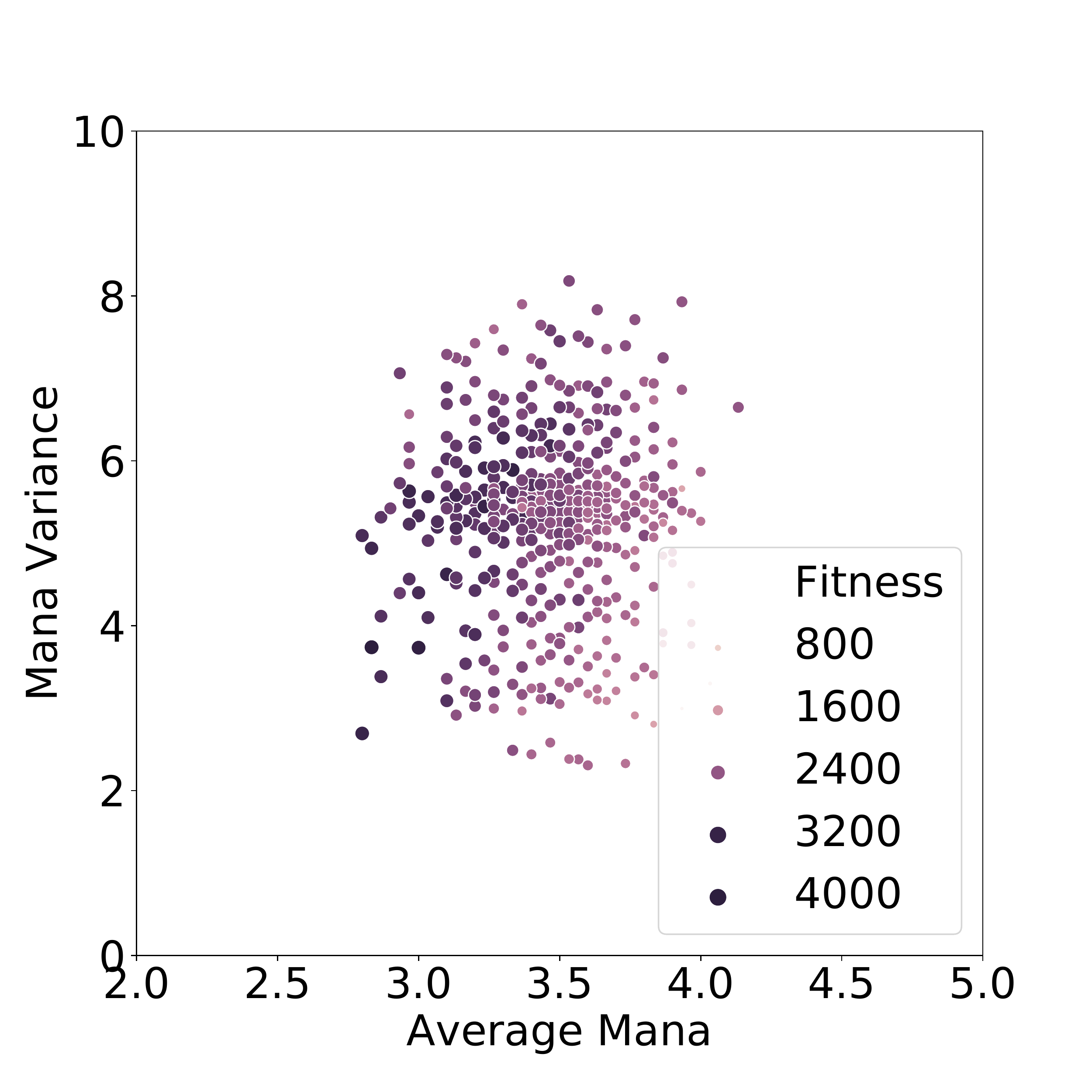}
    }
    \subfigure[Aggro Warlock]
    {
        \includegraphics[width=.32\textwidth]{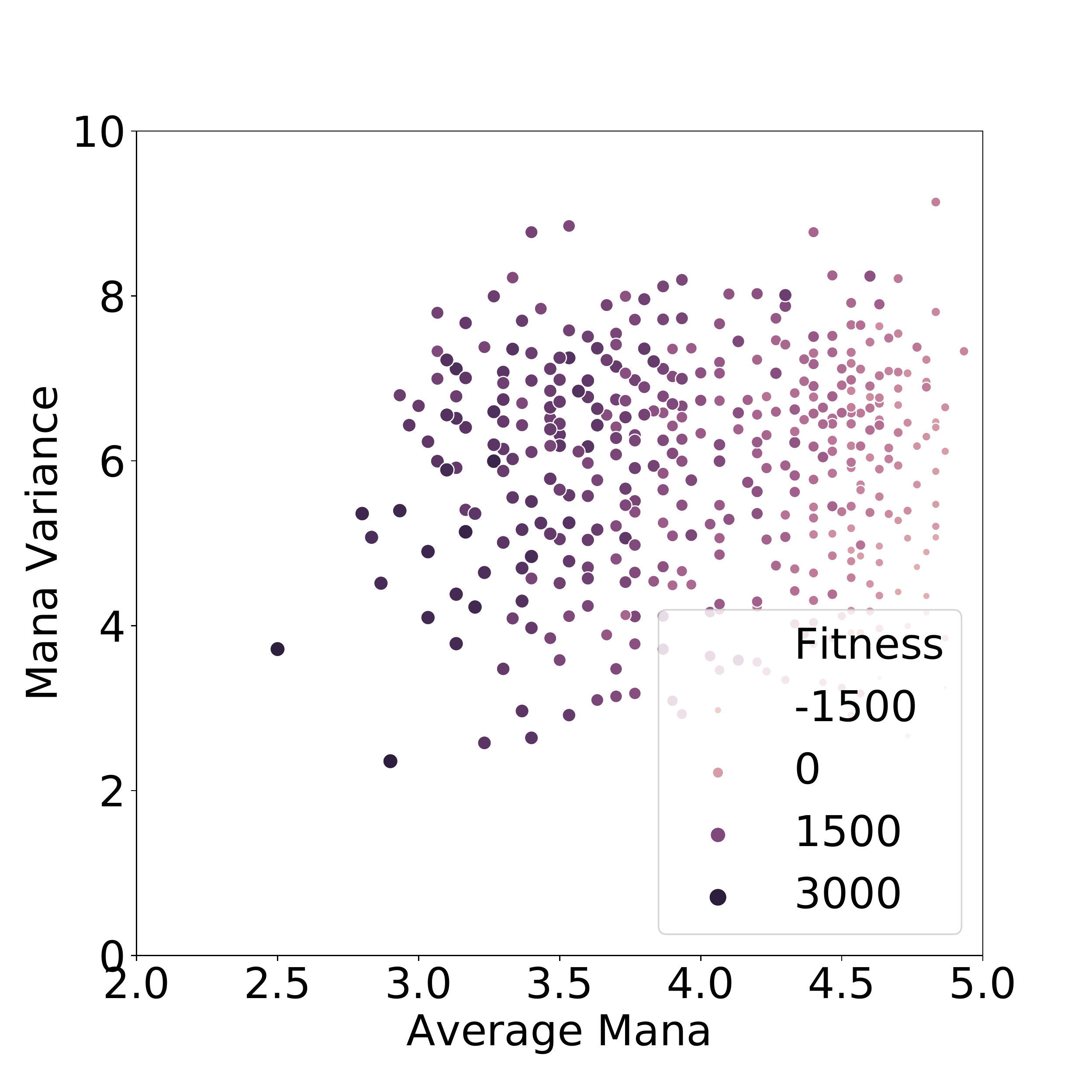}
    }
\caption{\emph{Distributions of Deck Performances Evolved Against Starter Decks in the MESB Validation Experiment.} 
\label{fig:expset1-scatterplot} Darker hues indicate positive performance (quantified by fitness). Scatterplots for different deck configurations and strategies exhibit different shapes. While the x and y axes are standard across each of the scatter plots, to better visualize patterns in archetypes hue refers to relative fitnesses described in the corresponding legends. Both the paladin and warlock aggro archetypes show stronger hero-relative decks where average mana cost is low. However, strong hero-relative control decks exist across a range of average mana costs. Interestingly, strong hero-relative decks for both aggro and control hunters exist when average mana cost is low, potentially indicating the that this archetype is nonviable.}
\end{figure*}
%Still, mana curve is an important and defining feature for players when they are building decks. To adapt mana curve to behavior features in MESB, experiments in this paper approximate curve through average mana and variation of the cards. 

To give an intuition for MESB's performance with respect to multiple play styles, Figure \ref{fig:expset1-scatterplot} shows the distribution of elite decks from the initial MESB validation experiment plotted in the space of mana curve space. 
%Results are from the first experiment evolved against the set of starter decks rather than results from the second experiment where decks were evolved against previously evolved opponents based on performance indicated in table \ref{tab:10k}.
%We chose to use the results of experiment set 1 in this case because the decks from that experiment appear to be stronger than in experiment set 2. 
Circles in figure \ref{fig:expset1-scatterplot} represent decks in the map of elites, while hue indicates fitness (summed difference in health at the end of each simulated game). Darker hues correspond to higher performance. The (x,y) coordinates in these plots represent position in approximated mana curve space. For example, in figure \ref{fig:expset1-scatterplot}a, control hunter decks with low average mana cost (toward the left of image) trend toward higher fitness than those with high cost (toward the right). Decks with lower variance and higher average mana cost (lower right) trend toward lower fitness, indicating that decks with only higher cost cards lose before they can play these powerful cards. 

Like aggro decks played by humans, trends in figures \ref{fig:expset1-scatterplot}d, \ref{fig:expset1-scatterplot}e, and \ref{fig:expset1-scatterplot}f show higher performance with low mana cost. If the decks have high mana cost, they can perform well when variance is high, indicating there are sufficiently many low-cost cards to be played early in the game. 
While high variance in cost can mitigate some impact of having a high average mana cost, in all three maps of elites played with the aggro strategy, high mana cost and low variance significantly impacts the ability of the player to execute a successful aggro strategy. Too many high cost cards leaves the player unable to defend or attack in early turns of the game.

In figures \ref{fig:expset1-scatterplot}b and \ref{fig:expset1-scatterplot}c, MESB discovers high performing decks for the control strategy across a spectrum of mana curves. Interestingly, the plot for the control hunter in figure \ref{fig:expset1-scatterplot}a is similar to the plot of the aggro hunter in figure \ref{fig:expset1-scatterplot}d. While the plot may indicate that hunter decks are naturally suited to aggro strategies, the aggro hunter loses to all of the other decks and gameplay strategies when the decks are compared (i.e. when they are played against each other 1000 times). In fact all of the control strategies win against their corresponding aggro strategies. While these plots are only shown for the three classes and two strategies in this experiment, class appears to impact the shape of these plots and is further discussed in section \ref{sec:discussion}.

\begin{figure}
\centering
\subfigure[Distribution of Decks in MESB Population]
{
\includegraphics[width=.22\textwidth]{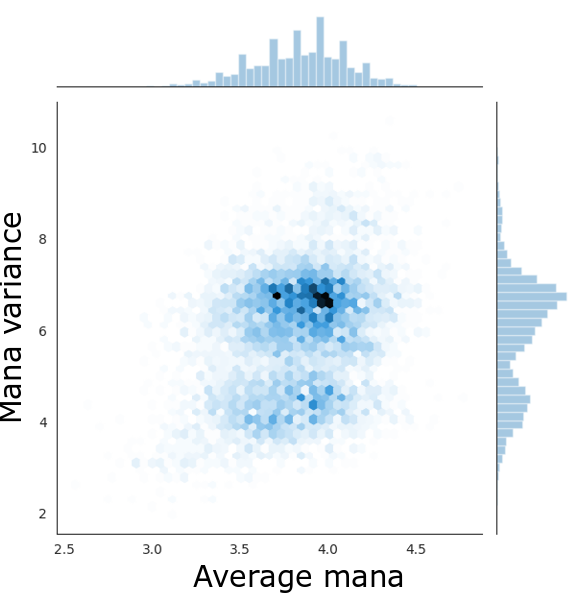}
}
\subfigure[Distribution of All Possible Decks in the Behavior Space]
{
\includegraphics[width=.22\textwidth]{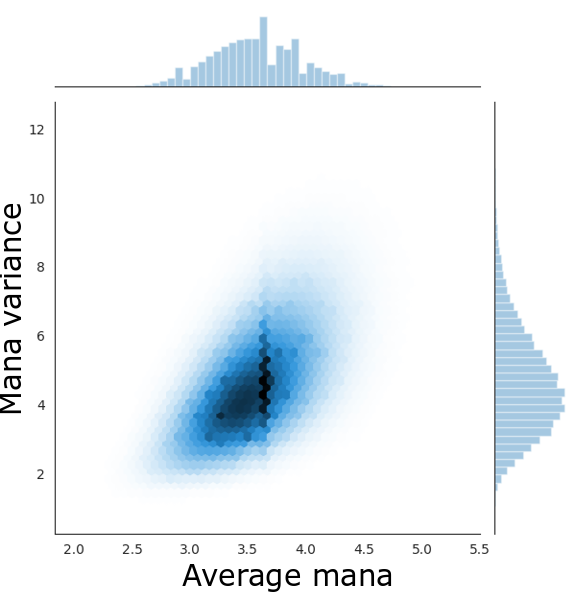}
}
%\subfigure[ME Behavior Range 1]
%{
%\includegraphics[width=.22\textwidth]{pala_feat_dist_me.png}
%}
%\subfigure[ME Behavior Range 2]
%{
%\includegraphics[width=.22\textwidth]{pala_feat_dist_me2.png}
%}
\caption{\emph{Density Distributions of Deck Populations.} Darker cells denote a higher density of decks. In (a) all 10,000 generated decks are plotted in behavior space. This map represents the control paladin decks in the elite adversaries experiment. The distribution of all possible decks is calculated and illustrated in the behavior space in (b). While MESB does oversample some regions of decks in the mana curve space, it is possible that those areas are where the highest performing decks are located. The ranges of average mana costs and variances in (a) matches those in the true distribution in (b). 
%MESB is especially appropriate when the range of values of the behavior characterizations is difficult to determine a priori (e.g. average turns per game, average damage per turn).
%The top-left figure shows distribution of all 10,000 decks seen when generating control paladin decks in the elite adversaries experiment. The top-right figure shows the distribution of all possible decks across the behavior space. The bottom-left and bottom-right figures show individual distributions of two runs of ME with different predefined grid boundary ranges and demonstrates a fundamental difficulty with applying ME to this problem.
\label{fig:curvedist}}
\end{figure}

To ensure that MESB effectively explored the mana curve space, figure~\ref{fig:curvedist} compares the mana curves discovered by MAP-Elites to the distribution of mana curves across the full deckspace. While the two distributions have different shape, the individuals observed by MESB span the majority of the mana curve space. 

%Dynamic programming helped calculate the number of decks with each possible value for average mana and mana variance.

% TODO figure below:
% Figure 7 should definitely be enlarged - the texts in the labels and legend are very hard or almost impossible to read.

\subsection{Performance of the Map of Elites}
%While traditional single-optimization approaches can produce high-quality Hearthstone decks \citep{bhatt2018exploring}, MESB has the potential to generate high quality decks exhibiting a variety of behaviors. Like traditional genetic algorithms it is important to show an increase in fitness over time, it is similarly important to show an increase in fitness of the decks in the map of elites.

\begin{figure*}
\centering
\includegraphics[width=\textwidth]{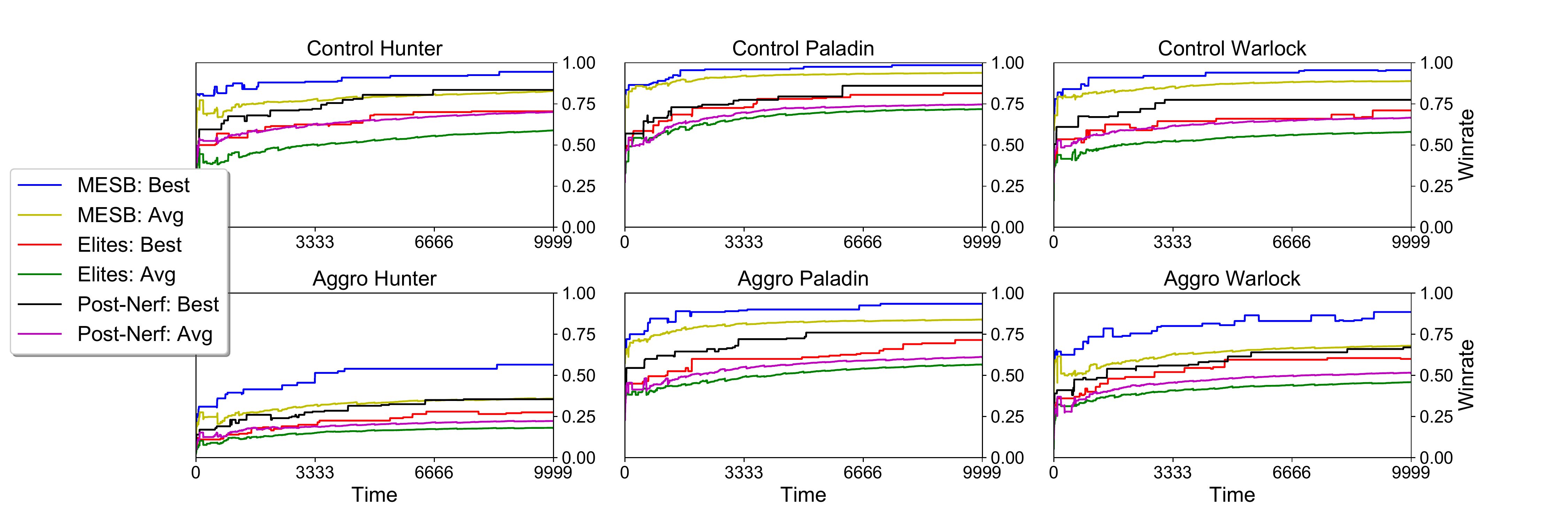}
\caption{\emph{Best and Average Winrates for MESB Validation, Elite Adversaries, and Deckbalancing Experiments.} The aggro hunter strategy generally performs worse than other deck configuration $\times$ strategy combinations in all experiments. In general, decks evolved in preliminary MESB validation experiments win more games than decks evolved against elite adversaries and decks with nerfed and buffed cards (weakened and strengthened, respectively, to balance the game). \label{fig:evograph123}}
\end{figure*}

For each experiment, figure \ref{fig:evograph123} shows the best and average winrates of the elite decks evolved with control and aggro strategies. Best and average winrates increase over time in all experiments. The aggro hunter decks in general perform significantly worse than other deck configuration $\times$ strategy combinations.

%it is important to show that MAP-Elites can be used to optimize along a provided evaluation metric over time, progressively generating higher quality decks over the course of a run. 

% TODO fig below
%Figure 5 should definitely be enlarged - the texts in the labels and legend are very hard or almost impossible to read.

% TODO paragraph below
% A bit more analysis on the plots from Figure 5 would be welcome. You focus mostly on 5a and 5b, but are the other plots not showing anything interesting? How's the relationship between experiments 1, 2, 3 in the 3 cases?

Figure \ref{fig:evograph123} shows that the best and average winrates in the MESB validation experiment were higher than those in the elite adversaries experiment. This result is expected given that opponents for evolved decks in the initial MESB validation experiment were less powerful than those faced by the evolved decks in the successive adversarial elites experiment, though it is important to note that lower winrate (against more challenging opponents) does not necessarily imply lower performance in general. In fact, when decks were compared against each other, the adversarial elites performed better than elites from their corresponding MESB validation. Generally, evolution should more easily identify high performing decks that evolved against weaker opponents because there are likely more combinations that perform well. Winrates for all deck configuration $\times$ strategy combinations in the elite adversaries experiment are lower than those in the MESB validation experiment, potentially suggesting that MESB initially found decks that won relatively consistently when evolving against weaker enemies. 
%Perhaps some of the best decks in the space were found through MESB in the initial validation experiment, thereby suggesting more transitivity in the space of possible card combinations than previously observed. It is additionally possible that the strength of the elite adversary decks was suppressing evolution and the number of iterations was insufficient to overcome the additional level of challenge.

\subsection{Measuring the Effect of Balance Changes}

\begin{figure}
    \centering
    \includegraphics[scale=0.7]{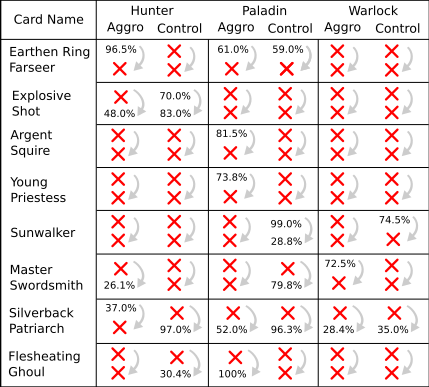}
    \caption{\emph{Changes in Card Frequencies after Nerfs and Buffs (Deckbalancing Experiment).} Cards with high frequency are \emph{nerfed} to weaken them and encourage exploration, whereas cards with low frequency may be \emph{buffed} to incentivize inclusion in decks. An $\times$ indicates occurrence in 25\% or fewer decks.
    \label{fig:cardChanges}}
\end{figure}

Results from the elite adversaries experiment suggest that even though high performing decks are distributed over a range of mana curves, decks often find specific cards critical to their performance. For example, as illustrated in figure \ref{fig:cardChanges}, Sunwalker is present in 99\% percent of the control paladin decks, Earthen Ring Farseer is in 96.5\% percent of aggro hunter decks, and Explosive Shot in 70\% of control hunter decks. While it is possible that these cards push search toward local optima, it is also possible that they are core elements of successful deck and strategy combinations. For the deckbalancing experiment, the value of one card shared by the majority of individuals in each archetype (e.g. control paladin, aggro warlock, etc.) is decreased to explore whether MAP-Elites can help evaluate the impact of card balancing on evolved decks. The mana cost of the Earthen Ring Farseer, Explosive Shot, Argent Squire, and Young Priestess are increased by two, the Sunwalker's attack decreased by three, and the Master Swordsmith's health decreased by two.

With the exception of two cards (Explosive Shot and Master Swordsmith), reducing the power of the remaining four cards (i.e. nerfing them) also reduces their presence in the map of elites shown in table \ref{fig:cardChanges}. Three of the cards are only found in 25\% or less of decks in the map of elites for each archetypes. The fourth (the Sunwalker card for control paladin) is reduced to 28.8\%. Interestingly, increasing the attack power by two increases the presence of the Silverback Patriarch above 25\% for five of the six deck configurations. Before reducing the power of cards, most of the cards were present in only one or two different class and gameplay strategy archetypes.

Similarly, nerfing Explosive Shot and Master Swordsmith increased their occurrence in aggro and control hunter decks and control hunter and aggro paladin decks, respectively. Explosive Shot is a spell card that costs five mana to play, and does five damage to a minion and two damage to its immediate neighbors. The nerf for Explosive Shot is an increase mana cost of two. While decks playing this card may benefit from a slightly different mana distribution due to less competition in a MAP-Elites cell, it is also possible that the additional cost forces the greedy, short sighted strategies to play the card later in the game when there are potentially more cards on the board and advantage to gain. This deckbalancing experiment demonstrates how MAP-Elites can potentially uncover complex relationships in gameplay (e.g. that it is a benefit to the card holder to force a card to be played later in the game).

%\begin{figure*}
%\centering
%\subfigure[Distribution of All Decks]
%{
%\includegraphics[scale=.4]{nerfed_paladin_control.png}
%}
%\subfigure[Distribution of Decks with Silverback Patriarch]
%{
%    \includegraphics[scale=.4]{silverbackpatriarch_nerfed_paladin_control.png}
%}
%\subfigure[Distribution of Decks with Sunwalker]
%{
%\includegraphics[scale=.4]{sunwalker_nerfed_paladin_control.png}
%}
%\caption{\emph{Distributions of Control Paladin Decks Post-Nerf}
%\label{fig:sunwalker-silverback}}
%\end{figure*}

The occurrence of the Silverback Patriarch varies between class and strategy archetypes. While originally it is only present in aggro hunter deck, after the buff it is present in all \emph{but} the aggro hunter decks. Because the card costs the same amount of mana in both the experiments, it is unclear what caused an aggro deck to abandon a taunt card with higher attack power. Perhaps more runs of map elites are necessary to make claims about individual cards. However, MAP-Elites still suggests interesting trends about the Silverback Patriarch in other decks. For example with the control paladin, the buffed Silverback Patriarch is  popular in cluster of decks with higher variance and lower cost. Among the same set of post-nerf control paladin decks, the Sunwalker's reduction in attack points pushed it in a seemingly unrelated direction to  decks with a higher average mana cost. Such effects can be difficult to predict or even detect in complex systems and their discovery here indicates a preliminary effectiveness of the MAP-Elites algorithm for game testing, balance, and design.

%The fact that the card is not universally used across all "players" is interesting as it indicates that the card was only buffed enough to be useful to certain playstyles. 
%This distribution is generally consistent with expectations. It naturally stands to reason that control players would favor a taunt card with reasonable health and enough power to force painful trades. 

%For 5 out of the 7 cards, the balance changes had expected results. The prevalence of overused cards was decreased drastically, and the usage of silverback patriarch increased dramatically in several class/card combinations. It is worth noting that the usage of silverback patriarch varies wildly based on class/strategy combination, indicating the strong influence of compatibility between the card and the player. The fact that the card is not universally used across all "players" is interesting as it indicates that the card was only buffed enough to be useful to certain playstyles. This distribution is generally consistent with expectations. It naturally stands to reason that control players would favor a taunt card with reasonable health and enough power to force painful trades. 

As a result of rebalancing the cards (i.e. the nerfs and buffs) and rerunning MAP-Elites, successful aggro paladin decks included at least one Flesheating Ghoul card in 100\% of decks. One hypothesis is that by increasing the cost of several cards a once, fewer of the low-cost minions that aggro paladins rely on were available for deckbuilding. Flesheating Ghoul likely filled the vacuum left by the more expensive nerfed Young Priestess. Identifying how card rebalances affect the performance of specific decks (such as those made by the Hearthstone community) is an area for future work.

\section{Discussion} \label{sec:discussion}

Results from all of the experiments illustrate that it is possible to generate high-performing decks across a range of mana distributions for a variety of playstyles. While adding only one extra card set (classic cards) than the classic optimization approach described by \citet{bhatt2018exploring}, each additional set can introduce new balances and imbalances that could make it difficult to evolve a variety of decks. Transitivity is a property that designers need to balance well to ensure good game play, and often cards are changed after their release to encourage such variety. However, that MESB can successfully find this range of strategies corroborates the usefulness of the algorithm as a tool for deck space analysis.

By examining the distribution of high-performing decks on the map of elites, it appears that certain gameplay strategies require particular mana distributions. For example, for each of the three classes playing an aggro strategy, the highest performing decks had low mana average. While intuitive, it is conceivable and likely that some high-cost cards could benefit an aggro playstyle. Interestingly, the correlation between low mana and good aggro decks is stronger for hunter and warlock than paladin. Again, Map-Elites helps provide a non-trivial insight into the design of the game and its cards.

%* Design space looking at different behaviors

%* While these plots are only shown for the three classes and two strategies in the experiments, class appears to impact the shape of these plots and worth further investigation in future work.

%* While the map of elites Control Hunter Distrbutuon looks like aggro hunter distribution

%*Nerfed Explosive shot showed up more in some decks, probably because of less 

The balance change experiments demonstrate how MESB can support game design. Potentially overpowered cards and card combinations were identfied with an Apriori analysis of evolved decks; the diversity assured by MESB implies that if a pattern is discovered in multiple decks, it is almost certainly powerful in a variety of settings (perhaps too powerful). After nerfing cards from these sets, new runs explored the impact on the distributions of cards. With the exception of the  Silverback Patriarch that was present in more high performing decks after a nerf, most nerfed cards were included in fewer high performing decks. Likely this increase is an artifact of characterizing the space of decks by mana curve. While the higher-cost version of this card is not inherently better than the lower-cost version, MESB its higher cost may place it in a less competitive position in behavioral space. 
%uniqueness could have to do with the higher-cost version of the card fitting into new mana distributions.

The main critique that could be leveled against the methodology employed in this paper is that it is dependent on a particular game-playing agent, namely the agent that comes with Sabberstone (the Hearthstone simulator used in these experiments). While it is unavoidable that any agent has a particular playstyle and will bias toward certain deck builds, the advantage of playing games with the Sabberstone agent is that it is well-tested and reasonably fast. Future work will explore how bias in these agents when compared to human playstyles. However, the same agaent plays games in all of the experiments; its parameterization between aggro and control styles is varied to mitigate playstyle bias.

Future will explore MESB and other modifications to MAP-Elites to investigate different problems in the space of Hearthstone decks. One question is whether different behavior vectors produce higher performing decks when played with a variety of agent strategies (beyond the aggro and control strategies in this paper). Such experiments could facilitate understanding of agent playstyles and preferences, or the formulation of more interesting behavior vectors to represent the decks. One particularly interesting avenue would be to attempt targeted rebalancing, to intentionally cultivate a metagame that favors a specific class or strategy. This methodology could help validate that balance changes affect the spaces in meaningful ways. Alternatively, it would also be interesting to observe the impact of new cards on the metagame, or even the evolution of new cards so as to occupy an unclaimed part of deck space.
 
% Discussion

% Did the results match our expectations?

% Discuss "strategic alignment" w.r.t. decks

% >> Lisa comment: Interesting that we are evolving a tool for humans (players) to use, but the algorithm used does not require a human evaluator in the loop. 

% How should we interpret the results?

% Highlight how the results are different from (and hopefully an improvement on) the previous Hearthstone paper 

% What have we learned? Are there any general lessons to take away?

% Demonstrated applicability to a massively popular game (real world impact)

% Potential topic: The role of evolution as an exploratory algorithm

% Potential topic: The role of evolution in games & content generation writ broadly

% Future 
% >> Note that future work could investigate MAP-Elites variants such as CVT-MAP-Elites

% >> Current work also doesn't take metagame into account

%%%%%%%AMY NOTES: 
%Fitness function design
%No incentive to use your hero power as a warlock (probably should record hero power usage)
%Hunters are not running as many secrets as it should be
%%%%%%

% TODO conclusion
% Discussion and Conclusion focus on how the proposed approach achieved "balance" and "high-performing" decks. However, the Introduction promised "a toolbox (i.e. a deck) around which a human player can construct winning strategies", which was never mentioned in the paper again.
% Note: copied from discussion

\section{Conclusion}
This paper explored deckbuilding in the game Hearthstone through a novel modification of the MAP-Elites algorithm that introduced sliding grid cell boundaries (MESB). A series of experiments revealed that MESB is able to discover high-performing decks in a variety of strategy spaces in addition to revealing potentially novel gameplay relationships. Not only do these results offer practical implications for improving playability of a massively popular real-world game, they also expand the reach of quality diversity algorithms beyond evolutionary robotics and into a new type of domain that can inform our theoretical understanding of this promising new class of algorithms.

%\begin{acks}
%Acknowledgements go here

%\end{acks}

\bibliographystyle{ACM-Reference-Format}
\bibliography{bibliography} 

\end{document}